\pdfoutput=1

\documentclass[11pt]{article}

\usepackage{EACL2023}

\usepackage{times}
\usepackage{latexsym}

\usepackage[T1]{fontenc}

\usepackage[utf8]{inputenc}

\usepackage{microtype}
\usepackage{graphicx}
\usepackage{amsmath}
\usepackage{amsthm}
\usepackage{booktabs}
\usepackage{algorithm}
\usepackage{algorithmic}
\usepackage{mathtools}
\usepackage{booktabs}
\usepackage{multirow}
\usepackage{microtype}
\usepackage{wasysym}
\usepackage{amssymb}
\usepackage{xspace}

\usepackage{inconsolata}

\usepackage{csquotes}
\usepackage[capitalise,noabbrev]{cleveref}
\usepackage{multirow}

\newcommand{\proposed}{}
\def\proposed/{\textit{SwitchPrompt}}

\newcommand{\TREC}{\textsc{Trec}\xspace}
\newcommand{\GARD}{\textsc{Gard}\xspace}
\newcommand{\SOFC}{\textsc{Sofc-E}xp\xspace}

\makeatletter
\renewcommand\paragraph{%
  \@startsection{paragraph}
    {4}
    {\z@}
    {3.25ex \@plus1ex \@minus.2ex}
    {-1em}
    {\normalfont\normalsize\bfseries\maybe@addperiod}%
}
\newcommand{\maybe@addperiod}[1]{%
  #1\@addpunct{.}%
}
\makeatother

\def\smallcol{\hspace{6pt}}
\def\tinycol{\hspace{4pt}}

\title{SwitchPrompt: Learning Domain-Specific Gated Soft Prompts\\
for Classification in Low-Resource Domains}

\author{
Koustava Goswami$^{1,2}$\thanks{$^*$Research work conducted during internship at Bosch Center for Artificial Intelligence. Contact: koustavag@adobe.com}~~, Lukas Lange$^2$, Jun Araki$^3$, \\ {\bf Heike Adel}$^2$ \\
$^1$ Adobe Research Bangalore, India \\
$^2$  Bosch Center for Artificial Intelligence, Renningen, Germany\\
$^3$ Bosch Research North America\\
{\tt \small koustavag@adobe.com},
{\tt \small \{Lukas.Lange,Heike.Adel\}@de.bosch.com},
{\tt \small Jun.Araki@us.bosch.com}
}

\begin{document}
\maketitle

\begin{abstract}
Prompting pre-trained language models leads to promising results across natural language processing tasks but is less effective when applied in low-resource domains, due to the domain gap between the pre-training data and the downstream task. In this work, we bridge this gap with a novel and lightweight prompting methodology called \textit{\proposed/}
for the adaptation of language models trained on datasets from the general domain to diverse low-resource domains. Using domain-specific keywords with a trainable gated prompt, \textit{\proposed/} offers domain-oriented prompting, that is, effective guidance on the target domains for general-domain language models. Our few-shot experiments on three text classification benchmarks demonstrate the efficacy of the general-domain pre-trained language models when used with \textit{\proposed/}. They often even outperform their domain-specific counterparts trained with baseline state-of-the-art prompting methods by up to $10.7\%$ performance increase in accuracy. This result indicates that \textit{\proposed/} effectively reduces the need for domain-specific language model pre-training.
\end{abstract}

\section{Introduction}\label{sec:intro}
Recent work showed promising results on different natural language processing tasks when prompting pre-trained language models (LMs) instead of fine-tuning them, especially in low-resource settings \cite{schucher-etal-2022-power}. 
Most LMs which are publicly available have been trained on general-domain corpora \cite{devlin-etal-2019-bert,liu1907roberta,goyal-etal-2021-larger}, such as Wikipedia or the BooksCorpus \cite{DBLP:conf/iccv/ZhuKZSUTF15}.
Applying them to tasks from a special domain results in a domain gap.

For some special domains, domain-specific LMs exists, e.g., Clinical BERT \cite{alsentzer-etal-2019-publicly} or BioBERT \cite{lee2020biobert}.
However, pre-training deep language models requires large amounts of text data.\footnote{Clinical BERT \cite{alsentzer-etal-2019-publicly}, for instance, was trained on the MIMIC-III $v1.4$ database \cite{johnson2016mimic} which includes $2$ million notes.}
While we can assume the availability of large-scale text data in the general domain, this assumption might not hold for
low-resource domains, making the creation of domain-specific LMs challenging. Moreover, training different models for each and every new domain might be inefficient
from a computation point of view.\footnote{\citet{lee2020biobert} used eight NVIDIA $V100$ GPUs for $23$ days to train the BioBERT.}
Even if there are domain-specific texts and computational resources available, domain-specific LMs may not be able to get sufficient domain-oriented guidance through traditional prompting techniques because, for instance, domain-specific knowledge might be represented by a large and diverse vocabulary. As a result, both prompting LMs from the general domain and from a special domain might be ineffective, especially in low-resource settings.


Motivated by these challenges, we explore domain-oriented prompts and propose a novel and lightweight method, \textit{\proposed/}, to effectively retrieve domain-specific knowledge from pre-trained LMs.
It extends the sequence of soft-prompting vectors with a sequence of vectors representing domain-specific keywords and introduces gates to allow the model to dynamically switch between a general soft prompt and a domain-specific
one
based on the input sentence.
We hypothesize that this approach helps to effectively retrieve domain-specific knowledge from pre-trained LMs.

Our experiments on benchmark datasets from different domains indicate that \proposed/ outperforms different state-of-the-art prompting methods. 
It improves results in both in-domain and out-of-domain settings, effectively reducing domain gaps among pre-training and downstream task data.
We find that it is especially suitable for low-resource settings (both little data and little computational resources) as it neither requires pre-training domain-specific LMs nor fine-tuning LMs for the downstream task. The code base for \textit{\proposed/} is available online.\footnote{\url{https://github.com/boschresearch/switchprompt}}

\section{Related Work}\label{sec:related}

\paragraph{Language model prompting}
Prompting pre-trained LMs has been shown effective for different NLP tasks~\cite{Brown2020Language}. 
While discrete prompts are intuitively understandable, their design requires non-trivial human involvement and they may be outperformed by fine-tuning~\cite{shin-etal-2020-autoprompt,jiang-etal-2020-know,gao-etal-2021-making}.  Recent studies address this issue by optimizing so-called soft prompts in continuous space.
\citet{li-liang-2021-prefix} propose prefix tuning that optimizes prefix activations prepended to the input layer and each layer in the encoder stack.
\citet{lester-etal-2021-power} prepend trainable continuous embeddings to the original sequence of input word embeddings. \citet{DBLP:journals/corr/abs-2103-10385} propose P-tuning in which an LSTM encoder 
captures the sequential representations of the soft prompts. 
\citet{liu-etal-2022-p} use a deep prompting methodology which injects prompts at each layer of the pre-trained LM. 
In contrast to those prior works, we propose a new soft prompting method that is especially suited for low-resource domains.

\paragraph{Language models in special
domains}
Most popular pre-trained LMs are trained on data from the general domain.
Tailoring an LM towards a domain can be done via domain-specific pre-training from scratch \citep[i.a.,][]{alsentzer-etal-2019-publicly,lee2020biobert} or adaptation of an existing model to target domain data with continued pre-training \citep[i.a.,][]{gururangan-etal-2020-dont,xu-etal-2020-dombert,lange-etal-2022-clinx}. We refer to the survey of 
\citet{hedderich-etal-2021-survey} for more information on language model adaptation for low-resource domains. 
In this paper, we take a different approach and investigate to which extend LMs that were pre-trained on the general domain can be prompted for domain knowledge in few-shot settings as this requires only a minimal amount of domain-specific data.

\section{Method}\label{sec:method}

We now present \proposed/, and give an example of an architecture in which it can be applied.  In our architectural setup, the underlying pre-trained language model is fixed (i.e., not fine-tuned). 

\subsection{Domain-Specific Soft Prompts}\label{sec:soft_prompts}
The motivation behind our proposed prompts $P \in \mathbb{R}^{l \times e}$ is to allow the model to dynamically switch between a general-domain prompt 
$P_g$
and a domain-specific prompt 
$P_d$
in order to retrieve different kinds of knowledge from the pre-trained LM based on the current input, where $e$ is the embedding dimension, and $l$ is the length of the prompt (i.e., number of soft-prompt vectors). 
We implement this with a sigmoid-based gating function:
\begin{align} \label{number1}
    P &= g_1 (\mathsf{pad}(P_g)) + (1 - g_1)P_d\\
    g_1 &= \sigma (w_1^\top s_{\text{input}})
\end{align}
where $\mathsf{pad}$ is a 
function that pads $P_g$ to length $l$. The prompts $P_g$ and $P_d$ will be defined in Equations \ref{general-prompt} and \ref{eq:domain-prompt}, respectively.
Gate $g_1$ is calculated based on the representation of the input sentence $s_{\text{input}}$ (in our case the representation of the $[\text{CLS}]$ token when feeding the input sentence into the pre-trained LM) and a weight vector $w_1 \in \mathbb{R}^{e}$ that is randomly initialized and updated during training.

The general-domain prompt is implemented as a sequence of randomly initialized vectors $v_1, \dots, v_m$ that are trained on the downstream task, similar to \newcite{lester-etal-2021-power} and \newcite{liu-etal-2022-p}:
\begin{equation} \label{general-prompt}
    P_g = [v_1; v_2; \dots; v_m] = V_m
\end{equation}
The sequence length $m$ is a hyperparameter of the model and `[;]' denotes concatenation.

The domain-specific prompt is designed to incorporate a sequence of vectors $K_n = [k_1; \dots; k_n$] that represent domain-specific keywords. The intuition is to inject the semantic information of the special domains using the domain-specific keywords.
We define the set of keywords using a term-frequency-based approach. 
In particular, we estimate a score $c$ for each word $w$ from the target domain based on the normalized term frequencies estimated on documents from the general domain $\text{tf}_g(w)$ and domain-specific documents $\text{tf}_d(w)$:
\begin{align}
    c(w) &= \alpha \cdot \text{tf}_g(w) + \text{tf}_d(w), \alpha < 0 \\
    t_i &= \{w | \text{rank}(c(w)) = i\}, 1 \le i \le n \label{eq:k} 
\end{align}

Using $\alpha < 0$, we are able to select terms that are representative for the target domain and avoid terms that are frequent in the general domain. In Equation \ref{eq:k}, we select the $n$ words from the target domain with the highest scores $c$ as our keywords. Each keyword $t_i$ is then represented as a vector $k_i$, using the same language model as for prompting.

Initial experiments showed that it is not enough to simply set $P_d = K_n = [k_1, k_2, \dots, k_n]$ but that the sequence of keywords should actually be combined with the sequence of soft prompts. Thus, we implement the domain-specific prompt as follows:
\begin{flalign}
\footnotesize
P_d &= g_2 [V_m;K_n] + (1 - g_2) [K_n;V_m] \label{eq:domain-prompt}\\
g_2 &= \sigma (w_2^\top s_{\text{input}})
\end{flalign}

We combine the sequence of keywords of length $n$ with the sequence of soft prompts of length $m$ with concatenation, yielding a sequence of length $l=m+n$. We let the model decide with a second gate $g_2$ in which order the sequences should be concatenated. Again, the gate is calculated based on the representation of the input sentence and a trainable weight vector $w_2 \in \mathbb{R}^{e}$, where $e$ is the embedding dimension. 
Thus, although the same domain-specific keywords are used for all inputs, the resulting soft prompt is dependent on the input sentence.

\subsection{Prompting Architecture} \label{prompting-architecture}
Since our proposed method is a new definition of soft prompts, it can be integrated into any existing model that uses soft
prompts. In our experiments, we adopt the \textit{P-Tuning v2} architecture \cite{liu-etal-2022-p} because of its high efficacy on different natural language understanding tasks.
P-Tuning v2 is an adaptation of deep prompt tuning \cite{DBLP:conf/naacl/QinE21,DBLP:conf/acl/LiL20} that injects soft prompts at every layer of the pre-trained LM. 
During training, the prompts are tuned but the LM stays fixed. For the class prediction of the downstream task, a randomly initialized classification head is added on top of the pre-trained LM.

\section{Experiments}
In this section, we describe the setup (datasets, training details and baselines) and the results of our experiments.

\subsection{Datasets}
For our experiments, we use classification benchmark datasets from different domains: question classification from the general domain \cite[\TREC,][]{voorhees-tice-2000-trec} and from the clinical domain \cite[\GARD,][]{kilicoglu-etal-2016-annotating}, as well as experiment classification from the materials science domain \cite[\SOFC,][]{friedrich-etal-2020-sofc}. 
Statistics of the datasets can be found in Table \ref{tab:statistics}.\begin{table}[h]
\centering
\small
\scriptsize
\resizebox{\columnwidth}{!}{%
	\begin{tabular}{llrr}
	\toprule
	{\bf Dataset} & {\bf Domain} & {\bf Instances} & {\bf Classes} \\
	\midrule
	\GARD & Clinical & 1253 & 11 \\
	\SOFC & Materials science & 2042 & 2 \\
	\TREC & General & 4893 & 7 \\
	\bottomrule
	\end{tabular}
}
	\caption{Statistics of text classification datasets.}
	\label{tab:statistics}
\end{table}

Among them, the \SOFC dataset offers a binary sentence classification task with positive and negative labels whereas \GARD and \TREC are multi-class question classification datasets.

To investigate very-low-resource settings, we construct \emph{few-shot datasets} by randomly sampling $N$ shots per class with $N \in \left\{2,4,16,64\right\}$.
Following the proposed theory for realistic \textit{low-resource regimes} \cite{DBLP:conf/nips/PerezKC21,kann-etal-2019-towards}, we also create \textit{few-shot development sets} by keeping the number of shots in the training and development sets in sync. In the $4$-shot scenario, for example, both the training and the development set consist of 4 examples for each class. 
For all datasets, we use accuracy $(\%)$ as evaluation metric. 

\begin{table*}[t]
\centering
\resizebox{\textwidth}{!}{%
	\begin{tabular}{lll}
	\toprule
	{\bf Domain} & {\bf Input} & {\bf Output} \\
	\midrule
	Clinical & How is it different from bilateral perisylvian polymicrogyria in how it presents ? & Diagnosis  \\
 Clinical & Are there products other than cigarette tobacco associated with Buerger disease ? & Cause  \\
	\hline
	Mat. science & They called this phenomenon nonfaradaic electrochemical modification of catalytic activity (NEMCA). & Negative  \\
 Mat. science & It is possible to reduce up to 35\% of NO present when the cell stacks are polarized with 1.5 V for each cell. & Positive  \\
	\hline
	General & What is the name of the largest city in Chile , South America ? & Location  \\
 General & What was the average life expectancy during the Stone Age ? & Number  \\
	\bottomrule
	\end{tabular}
}
	\caption{Example sentences and their labels from our domain-specific and general-domain datasets.}
	\label{tab:domain}
 \vspace{-0.2cm}
\end{table*}

To give a closer insight into the challenges of the different domains that we use in our experiments, we present example instances from the datasets in Table \ref{tab:domain}.
The examples show that the models need to cope with domain-specific terminology, such as \enquote{perisylvian polymicrogyria} (clinical domain) or \enquote{electrochemical} (materials science domain), and domain-specific labels, for instance, \enquote{diagnosis}.

\subsection{Training Details}
We use open-sourced HuggingFace language models\footnote{\url{https://huggingface.co/models}} for our experiments. 
 We train our models with a batch size of $32$. The maximum sequence length is set to $128$ and we use dropout with rate $0.1$ on the classification layer. We use the \textit{ExponentialLR}\footnote{\url{https://pytorch.org/docs/stable/optim.html\#torch.optim.Optimizer}} learning rate scheduler with a gamma value of $0.95$ and the Adam optimizer. All experiments are performed on a $V100$ GPU.\footnote{We ran our experiments on a carbon-neutral GPU cluster.} 
 Each reported result is the average performance of five runs.

 \subsection{Baselines}
We compare our method to different baselines: (i) Fine-tuning of the pre-trained LM, (ii) prompting using P-tuning \cite{DBLP:journals/corr/abs-2103-10385}, and (iii) prompting using P-Tuning v2 \cite{liu-etal-2022-p}.
For all methods, we report results for using either a general-domain LM (BERT \cite{devlin-etal-2019-bert}) or domain-specific LMs (Clinical BERT \cite{alsentzer-etal-2019-publicly} and SciBERT \cite{DBLP:conf/emnlp/BeltagyLC19}). 

\subsection{Results}

\paragraph{Low-resource domains.}
\begin{table}[t]
\centering
\def\tinycol{\hspace{4pt}}
\scriptsize
\begin{tabular}{@{} l@{\tinycol}p{1.5cm}@{\tinycol}p{0.75cm}@{\tinycol}p{0.75cm}@{\tinycol}p{0.85cm}@{\tinycol}p{0.87cm}@{\tinycol}p{0.3cm}@{\tinycol}}
\toprule
\textbf{Methodology} & \textbf{Model} & \textbf{2-shots} & \textbf{4-shots} & \textbf{16-shots} & \textbf{64-shots} & \textbf{All} \\ \midrule

\multirow{2}{*}{Fine-tuning}  & {BERT} & {21.2}          & {25.5}          & {40.8}          & {67.4}          & {81.8}            \\
 & {Clinical BERT} & {35.9}          & {40.4}          & {56.3}          & {68.1}          & {82.5} \\
\hline
\multirow{2}{*}{P-tuning}  & {BERT} & \textbf{48.3}          & {48.9}          & {53.1}          & {68.1}          & {82.0}            \\
 & {Clinical BERT} & \textbf{49.2}          & {53.1}          & {58.2}          & {69.6}          & {82.8}            \\
\hline
\multirow{2}{*}{P-tuning V2}  & {BERT} & {27.2}          & {44.4}          & {61.9}          & {79.1}          & {84.0}            \\
 & {Clinical BERT} & {34.3}          & {48.7}          & {63.4}          & \textbf{82.3}          & {86.7}            \\
\hline

\multirow{2}{*}{SwitchPrompt}  & {BERT} & {36.3}          & \textbf{54.2}          & \textbf{64.0}          & \textbf{81.1}          & \textbf{85.4}            \\
 & {Clinical BERT} & {40.9}          & \textbf{55.2}          & \textbf{65.1}          & {81.9}          & \textbf{86.9}            \\
\bottomrule
\end{tabular}
\caption{
Results on special-domain dataset \GARD.
}
\label{table:1}
\end{table}

\begin{table}[t]
\centering
\def\tinycol{\hspace{4pt}}
\scriptsize
\begin{tabular}{@{} l@{\tinycol}p{1.0cm}@{\tinycol}p{0.75cm}@{\tinycol}p{0.75cm}@{\tinycol}p{0.85cm}@{\tinycol}p{0.87cm}@{\tinycol}p{0.3cm}@{\tinycol}}
\toprule
\textbf{Methodology} & \textbf{Model} & \textbf{2-shots} & \textbf{4-shots} & \textbf{16-shots} & \textbf{64-shots} & \textbf{All} \\ \midrule

\multirow{2}{*}{Fine-tuning}  & {BERT} & {18.2}          & {26.1}          & {48.5}          & {54.6}          & {61.9}            \\
 & {SciBERT} & {29.4}          & {32.7}          & {50.4}          & {56.2}          & {64.7} \\
\hline
\multirow{2}{*}{P-tuning}  & {BERT} & \textbf{37.5}          & \textbf{38.2}          & {52.6}          & {58.5}          & {64.9} \\
 & {SciBERT} & \textbf{42.1}          & \textbf{43.4}          & {54.8}          & {59.3}          & {66.2}            \\
\hline
\multirow{2}{*}{P-tuning V2}  & {BERT} & {30.8}          & {31.2}          & {52.8}          & {59.9}          & {68.4}            \\
 & {SciBERT} & {33.7}          & {35.6}          & {53.9}          & {61.4}          & {69.7}      \\
\hline
\multirow{2}{*}{SwitchPrompt}  & {BERT} & {32.4}          & {34.3}          & \textbf{53.4}          & \textbf{61.0}          & \textbf{69.9}            \\
 & {SciBERT} & {36.2}          & {37.1}          & \textbf{55.9}          & \textbf{62.5}          & \textbf{70.6}            \\\bottomrule

\end{tabular}
\caption{
Results on special-domain dataset \SOFC.
}
\label{table:2}
\end{table}

Tables \ref{table:1} and \ref{table:2} show the results of our model for the clinical and materials science domain in comparison to state-of-the-art baseline approaches. In general, the prompting methods outperform fine-tuning, with especially large margins for very-few-shot settings ($2$ and $4$ shots). This highlights the limitations of fine-tuning with limited training datasets. Another general trend is that using domain-specific LMs (Clinical BERT and SciBERT, respectively) outperforms BERT from the general domain. Our proposed method \proposed/ outperforms other state-of-the-art prompting methods up to 2.1\% points. We further note that (i) our method prompting general-domain LMs even outperforms other methods prompting domain-specific LMs, and (ii) our method reduces the performance gap between using an LM from the general domain vs.\ a domain-specific one.


For very-few-shot settings (2,4-shots), P-tuning outperforms our method. We assume that the reason is that it replaces the input of the LM with differential embeddings from the prompt-encoder, while in our method we consider the vanilla inputs of the LM, reducing the complexity and training time (see Figure \ref{fig:analysis})  of our model. 
\paragraph{General domain.}
\begin{table}[t]
\centering
\def\tinycol{\hspace{4pt}}
\scriptsize
\begin{tabular}{@{} l@{\tinycol}p{0.8cm}@{\tinycol}p{0.75cm}@{\tinycol}p{0.75cm}@{\tinycol}p{0.85cm}@{\tinycol}p{0.87cm}@{\tinycol}p{0.3cm}@{\tinycol}}
\toprule
\textbf{Methodology} & \textbf{Model} & \textbf{2-shots} & \textbf{4-shots} & \textbf{16-shots} & \textbf{64-shots} & \textbf{All} \\ \midrule
{Fine-tuning}  & {BERT} & {33.3} & {53.3} & {71.4}          & {88.7} & {95.7}  \\
{P-tuning V2}  & {BERT} & {56.0} & {63.3} & {79.4}          & \textbf{{92.5}} & {96.8}  \\
{SwitchPrompt}  & {BERT}  & \textbf{{66.7}} & \textbf{{72.4}} & \textbf{{88.3}}          & {91.2} & \textbf{{97.6}}     \\
\bottomrule

\end{tabular}
\caption{
Results on general-domain dataset \TREC. 
}
\label{table:5}
\vspace{-0.3cm}
\end{table}

To investigate the behavior of our method in the general domain, we now evaluate its performance on \TREC. Table \ref{table:5} shows that our method outperforms both fine-tuning and other prompting methods in almost all dataset settings, up to 10.7 accuracy points. Thus, even on the general domain, \proposed/ can boost the performance of pre-trained LMs.

\begin{table*}[t]
\centering
\resizebox{\textwidth}{!}{%
	\begin{tabular}{ll}
	\toprule
	{\bf Domain} & {\bf Keywords} \\
	\midrule
	Clinical & Diagnosed, Prognosis, Cantrell, Idiopathic, Tourette, Opitz, Testotoxicosis, Late-onset, Amniocentesis, Prenatally  \\
	Mat. science & Fuel, Oxide, D8-Discover, Viscometer, Hydroxide/poly, Room-temperature, Ion-conductor, Electrocatalytic, Cobalt-doped, Non-homogeneous  \\
	General & Cholera, Tasman, Conservancy, Boil, Premier, Consumption, Conditioner, Foster, Chemiosmotic, Registers  \\
	\bottomrule
	\end{tabular}
}
	\caption{ Automatically selected $10$ keywords per domain by our approach.}
	\label{tab:keywords}
 \vspace{-0.2cm}
\end{table*}

\section{Analysis}
In this section, we report the results of our ablation study, give more insights into what our model learned and analyze its training time. We also provide a qualitative error analysis.

\begin{table}[t]
\centering
\def\tinycol{\hspace{3pt}}
\scriptsize
\begin{tabular}{@{}l@{\tinycol}l@{\tinycol}r@{}}
\toprule
& {\textbf{Prompt}} & {\textbf{Acc}}  \\ 
\midrule
(1) & $ g_1 (\mathsf{pad}(V_m)) + (1 - g_1) (g_2 [V_m;K_n] + (1 - g_2) [K_n;V_m] ) $ & \textbf{85.4} \\
(2) & $ g_1 (\mathsf{pad}(V_m)) + (1 - g_1) (g_2 V_m + (1 - g_2) K_n) $  & {82.6}  \\
(3) & $ g_1 (\mathsf{pad}(V_m)) + (1 - g_1) [V_m ; K_n] $  & {81.4}  \\
(4) & $ g_1 (\mathsf{pad}(V_m)) + (1 - g_1) [K_n ; V_m] $  & {77.6}  \\
(5) & $ K_n $  & {54.8}  \\
(6) & $ V_m $ & {84.0}      \\
\bottomrule
\end{tabular}
\caption{Impact of prompt design choices in the full-data setting of the \GARD dataset using BERT embeddings. Row (1) corresponds to \proposed/, and row (6) corresponds to P-Tuning v2.
}
\label{tab:ablation}
\end{table}

\paragraph{Ablation study.}

Our ablation study in Table \ref{tab:ablation} shows the impact of the different components of our prompting function,
evaluated with the full \GARD dataset.
Row (1)
corresponds to 
\proposed/, and
row (6)
corresponds to the previous state-of-the-art prompting model P-Tuning v2.
Row (2)
shows that the concatenation of keywords and the general-domain soft prompt is important to the model.
Row (3) and (4)
show the large impact of the second gate $g_2$, and
row (5) and (6)
show that neither the domain-specific keywords $K_n$ nor the general soft-prompting vectors $V_m$ alone are sufficient to achieve the highest performance.

\paragraph{Domain-specific keywords.} The keywords are an integral part of \proposed/. Since we compute keywords automatically (see Section \ref{sec:method}), we analyze the extracted keywords in more detail.


Table \ref{tab:keywords} shows the $10$ keywords that have been selected by our method. For the clinical and materials science domain, the keywords are domain-specific terms while for the general domain, the keywords cover a broad range of topics.

\begin{figure}[t]
    \centering
    \includegraphics[width=\columnwidth]{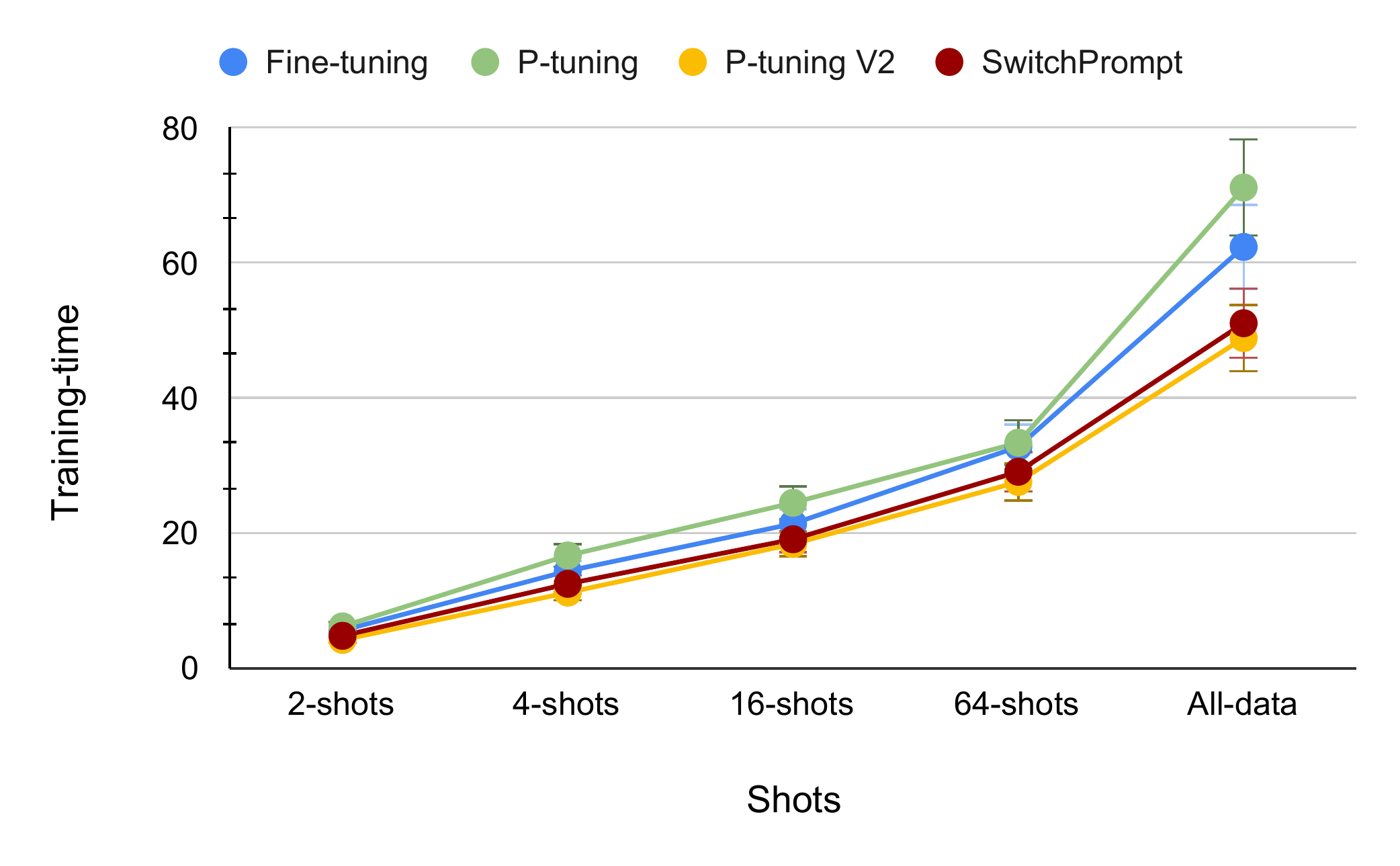}
    \caption{Training time in minutes for 10 epochs for different methods on the \GARD dataset.}
    \label{fig:analysis}
    \vspace{-0.5cm}
\end{figure}

\paragraph{Training time analysis.} During training time, the underlying LM is frozen in the \proposed/ framework. This substantially reduces training time and computational memory, compared to alternative approaches, such as fine-tuning or P-Tuning. Figure \ref{fig:analysis} illustrates this.
P-Tuning v2 is a little bit faster than our approach as it does not need to train the gating parameters. However, the time difference is considerably small ($2.2$ min for 10 epochs in the all-data setting, i.e., 0.22 min per epoch). 

\paragraph{Qualitative error analysis.} 


\begin{table}[t]
\centering
\footnotesize
\begin{tabular}{@{}p{3.8cm}@{\smallcol}l@{\smallcol}l@{}}
\toprule
{\bf Input} & {\bf Prediction} & {\bf Gold Output} \\
\midrule
\textit{How can this be?} & Management & Susceptibility \\
\textit{Will we be okay?} & Information & Prognosis \\
\textit{What is the treatment of mixed connective tissue disorder ?} & \multirow{2}{*}{Information} & \multirow{2}{*}{Management} \\
\textit{What are the expected outcomes for individuals with cryoglobulinemia ?} & \multirow{3}{*}{Information} & \multirow{3}{*}{Prognosis} \\
\bottomrule
\end{tabular}
\caption{Error analysis on GARD dataset.}
\label{tab:Error}
\vspace{-0.4cm}
\end{table}

Finally, we manually conduct a qualitative error analysis on the \GARD dataset. The results are displayed in Table~\ref{tab:Error}. We find that our method mainly fails when the input sentences convey little domain-specific information (see examples in first two rows). 
Another category of errors is the prediction of a more general class (\enquote{Information} instead of \enquote{Management} or \enquote{Prognosis} in the last two rows).



    

\section{Conclusion}

In this paper, we proposed a new methodology called \proposed/ for effectively prompting pre-trained language models in low-resource domains.
Integral parts of our method are domain-specific keywords and gates, which allow the language model to dynamically retrieve domain-specific knowledge.
Experiments on sentence classification datasets from different domains show that our method outperforms various baseline methods in few-shot and all-data settings. In particular, it reduces the performance gap between general-domain and domain-specific language models. Future work can investigate the impact on sequence-labeling tasks as well as on mixed-domain datasets.

\section*{Acknowledgments}
We would like to thank the members of the BCAI NLP \& NS-AI research group and the anonymous reviewers for their helpful comments.

\section*{Limitation}
In preliminary experiments, we found that our method is sensitive to the selection of keywords. While we found an automatic and domain-independent way for extracting them (see Section \ref{sec:method}), its efficacy needs to be tested on more domains and possibly also on mixed domain datasets.


\bibliography{anthology,custom}
\bibliographystyle{acl_natbib}

\end{document}